\documentclass[conference]{IEEEtran}

\usepackage{cite}
\usepackage{amsmath,amssymb,amsfonts}
\usepackage{graphicx}
\usepackage{booktabs}
\usepackage{multirow}
\usepackage{flafter}
\usepackage{textcomp}
\usepackage{xcolor}
\usepackage{url}
\usepackage{float}
\newcommand{\vect}[1]{\boldsymbol{#1}}

\begin{document}

\title{PACT: A Decoder-Confidence-Guided Two-Round Receiver for Joint Channel and Phase-Noise Estimation in Sub-THz OFDM}

\author{\IEEEauthorblockN{Ruilin Wang, Xiaodai Dong}
\IEEEauthorblockA{Department of Electrical and Computer Engineering, University of Victoria\\
Victoria, Canada\\
xdong@uvic.ca}}

\maketitle

\begin{abstract}
Frequency-selective fading and oscillator phase noise (PN) jointly impair coded sub-THz OFDM reception. We propose a decoder-confidence-guided two-round receiver combining a {pilot-aware complex transformer} (PACT) estimator, banded linear minimum mean-square error equalization, and neural normalized min-sum decoding. PACT jointly estimates the frequency-domain channel and effective time-domain PN through interacting complex-valued branches. Decoder-derived pseudo-pilots and continuous constellation-aware confidence guide second-round refinement through PACT inputs and attention, with parameters shared across rounds. At $E_b/N_0=30$~dB, the PACT receiver achieves approximately $70\%$ and $57\%$ lower code-block error rates than the best-performing learning-based baseline in the two benchmark settings, respectively. Frozen-weight tests further demonstrate applicability across channel/PN models, carrier frequencies, and FFT/bandwidth configurations.
\end{abstract}


\section{Introduction}
\label{sec:introduction}

Sub-terahertz bands around 100~GHz offer contiguous bandwidth
for high-rate 6G links~\cite{jiang2024terahertz},
but wideband multipath propagation {produces}
environment-dependent, frequency-selective
channels~\cite{3gpp38901}.
Oscillator phase noise (PN) introduces time-domain phase
fluctuations, causing common phase error (CPE) and
inter-carrier interference (ICI) in cyclic-prefix
orthogonal frequency-division multiplexing
(CP-OFDM)~\cite{armada2001understanding}.
{Modeling PN solely as additive white
Gaussian noise (AWGN) neglects its CPE
and ICI~\cite{petrovic2007effects},
motivating explicit joint estimation of the channel and PN.}
In coded reception, {both estimates affect
compensation, equalization, and decoding},
while decoder outputs can provide additional references
{for refinement}.
{These interactions motivate an integrated
receiver design.}

Classical receivers employ pilot-aided least-squares estimation,
model-based optimization, and extended Kalman
filtering~\cite{mehrpouyan2012joint,wang2017effective}.
Decision-directed and code-aided methods further reuse
detected symbols or decoder outputs to refine channel and PN
estimates~\cite{corvaja2009joint,wang2026iterative}.
Learning-based methods without channel coding include
staged channel and PN recovery~\cite{mohammadian2021deep,mattu2022learning},
single-pass joint channel and PN estimation with
CRFCAN~\cite{wang2026crfcan},
and SIRIUS for channel refinement using pseudo-pilots
selected by constellation-distance thresholds on equalized
symbols rather than decoder posteriors~\cite{shi2026unsupervised}.
Coded receivers include DeepSRX for soft-bit
prediction~\cite{marasinghe2025phase},
JCDDNet for joint channel estimation, detection, and
decoding without explicit PN modeling~\cite{sun2024trainable},
{and a DL-based JCDD receiver for
decoder-confidence-guided channel refinement with
superimposed pilots~\cite{li2026learning}.}

{These advances motivate combining
learning-based modeling of the statistical structure
of coupled channel and PN distortions
\cite{mohammadian2021deep,mattu2022learning}
with reliability-guided iterative processing.
Decoder outputs provide references beyond known pilots,
while continuous confidence conveys their reliability
to the network.
Reusing the same neural estimator enables further joint
channel and PN refinement without storing multiple sets
of estimator parameters.
Therefore, we propose the two-round receiver,
combining a pilot-aware complex transformer (PACT)
estimator, banded linear minimum mean-square error
(LMMSE) equalization, and neural normalized min-sum
(NNMS) decoding.}
The contributions are summarized as follows:
\begin{itemize}
    \item A dual-domain complex-valued PACT estimator
    jointly estimates the frequency-domain channel response
    and time-domain PN through interacting branches,
    with parameters shared across the two receiver rounds.

    \item Continuous pseudo-pilot confidence is derived from
decoder posteriors and constellation geometry.
It guides joint channel and PN refinement through
inputs and attention within PACT.
\end{itemize}
{The resulting PACT receiver achieves approximately
$70\%$ and $57\%$ lower code-block error rates than the
best-performing learning-based baseline at $E_b/N_0=30$~dB
in the two benchmark settings, respectively.}

\begin{figure*}[t]
    \centering
    \includegraphics[width=0.98\textwidth]{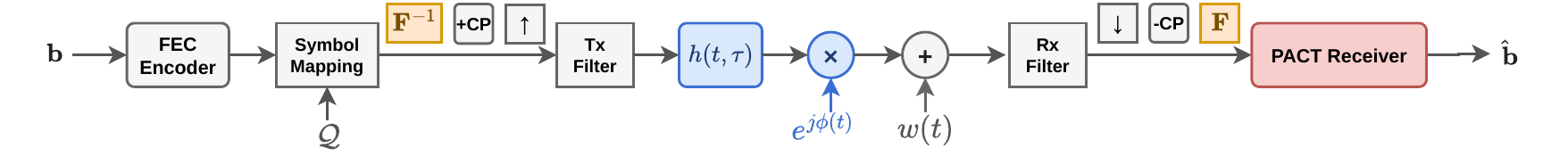}
    \caption{Coded sub-THz OFDM system with the proposed receiver.}
    \label{fig:system_model}
    \vspace{4pt}
\end{figure*}

\section{System Model}
\label{sec:system_model}

\subsection{Coded OFDM Signal Model}
\label{subsec:ofdm_model}

We consider the coded single-antenna OFDM system in
Fig.~\ref{fig:system_model}, with $N$ subcarriers and $M$
OFDM symbols per frame.
Information bits {are encoded using a low-density
parity-check (LDPC) code}, rate-matched to code rate $R_c$,
and Gray-mapped onto a unit-average-energy
constellation $\mathcal{Q}$.
The frequency-domain transmit grid $\boldsymbol{X}=[X_{k,m}]$
is assembled from data symbols and known unit-energy QPSK
pilots on the disjoint data and pilot subcarrier sets
$\mathcal{D}_m$ and $\mathcal{P}_m$ of the $m$th OFDM symbol,
respectively.
After an $N$-point IDFT, the OFDM symbols are each
extended with a CP of $N_{\rm cp}$ samples and
serialized into $s[i]$.
With upsampling factor $U$ and transmit filter $g_{\rm tx}$,
the transmit waveform $u[q]$, noiseless channel output
$v[q]$, and receive-filter input $r[q]$ are
\begin{align}
    u[q]
    &= \sum_i s[i]g_{\rm tx}[q-Ui], \nonumber\\
    v[q]
    &= \sum_{\ell=0}^{L-1}h[q,\ell]u[q-\ell], \nonumber\\
    r[q]
    &= e^{j\phi[q]}v[q]+w[q],
    \label{eq:ofdm_highrate}
\end{align}
where $i$ indexes samples before upsampling and $q$
indexes samples after upsampling.
Here, $h[q,\ell]$ is the length-$L$ time-varying
propagation channel, $\phi[q]$ is the PN process applied
before receive filtering, and
$w[q]\sim\mathcal{CN}(0,\sigma_w^2)$ is white Gaussian noise.
After matched filtering, ideal timing synchronization,
downsampling by $U$, and CP removal, the received samples
and their DFT are
\begin{align}
    y_m[n]
    &= (g_{\rm rx}*r)[q_{m,n}], \nonumber\\
    Y_{k,m}
    &= \sum_{n=0}^{N-1}y_m[n]e^{-j2\pi kn/N},
    \label{eq:ofdm_chain}
\end{align}
for $0\le k,n<N$ and $0\le m<M$, where
$q_{m,n}=d+U(i_m+N_{\rm cp}+n)$ is the high-rate index
of the $n$th retained sample of symbol $m$.
The receive filter $g_{\rm rx}$ is matched to $g_{\rm tx}$,
$i_m$ is the starting index of the $m$th CP-extended symbol
before upsampling, and $d$ compensates the combined
transmit/receive filter delay.

\subsection{Channel and Phase Noise Models}
\label{subsec:channel_pn_model}

The propagation channel $h[q,\ell]$ comprises
$N_{\rm path}$ paths with delays $\tau_p$ and time-varying
complex gains $\alpha_p(t)$ accounting for Doppler effects.
The channel estimation target is defined at $t_m$, the
midpoint of the $m$th OFDM symbol excluding its CP:
\begin{align}
    h_m^{\rm eq}[\ell]
    &= \sum_{p=1}^{N_{\rm path}}\alpha_p(t_m)\,
       g\!\left(\ell T_{\rm s}-\tau_p\right), \nonumber\\
    H_{k,m}
    &= \sum_{\ell\in\mathcal{T}_h}h_m^{\rm eq}[\ell]\,
       e^{-j2\pi k\ell/N},
    \label{eq:channel_target}
\end{align}
where $g(t)$ is the combined response of $g_{\rm tx}$ and
$g_{\rm rx}$ after delay compensation, $T_{\rm s}$ is the
baseband sampling interval, and $\mathcal{T}_h$ is the set
of retained channel-tap indices, which may include negative
indices after delay compensation.
The PN process is generated as
$\phi[q]=(g_\phi*\xi)[q]$, where
$\xi[q]\overset{\mathrm{i.i.d.}}{\sim}\mathcal{N}(0,1)$
and $g_\phi$ is a shaping filter designed to match the
selected PN PSD.
After receive processing, the effective PN target is
\begin{equation}
    p_m[n]
    = \frac{z_m^{(\phi)}[n]}{z_m^{(0)}[n]}
    = \frac{
        \sum_\nu g_{\rm rx}[\nu]v[q_{m,n}-\nu]
        e^{j\phi[q_{m,n}-\nu]}
    }{
        \sum_\nu g_{\rm rx}[\nu]v[q_{m,n}-\nu]
    },
    \label{eq:effective_pn}
\end{equation}
for $z_m^{(0)}[n]\ne0$.
Here, $z_m^{(\phi)}[n]$ and $z_m^{(0)}[n]$ are noiseless
samples obtained by the receive processing
in~\eqref{eq:ofdm_chain} with and without PN, respectively.
The receive filter $g_{\rm rx}$ combines samples with
different phases, so $p_m[n]$ may differ from
$e^{j\phi[q_{m,n}]}$ in both magnitude and phase.

The frequency-domain received signal can be expressed as
\begin{equation}
\begin{aligned}
    Y_{k,m}
    &= P_{0,m}H_{k,m}X_{k,m}\\
    &\quad+\sum_{\kappa=1}^{N-1}P_{\kappa,m}
      H_{k-\kappa,m}X_{k-\kappa,m}+W_{k,m},
\end{aligned}
\label{eq:pn_cpe_ici}
\end{equation}
where
$P_{\kappa,m}=(1/N)\sum_{n=0}^{N-1}
p_m[n]e^{-j2\pi\kappa n/N}$, and $\kappa$ denotes the subcarrier offset for
PN-induced interference from subcarrier $k-\kappa$ to subcarrier $k$.
For $k-\kappa<0$, the source subcarrier index is $k-\kappa+N$.
The term $W_{k,m}$ collects the transformed receive noise
and residual interference caused by within-symbol channel
variation and taps outside the CP-protected window.
The phase of the common coefficient $P_{0,m}$ gives CPE,
while the remaining terms describe ICI.
The targets $\vect{H}=[H_{k,m}]$ and $\vect{p}=[p_m[n]]$
are $N\times M$ grids in frequency and time, respectively.

\section{Proposed Receiver}
\label{sec:pact}

\begin{figure*}[t]
    \centering
    \setlength{\abovecaptionskip}{3pt}
    \includegraphics[width=0.98\textwidth,trim=0 6bp 0 10bp,clip]{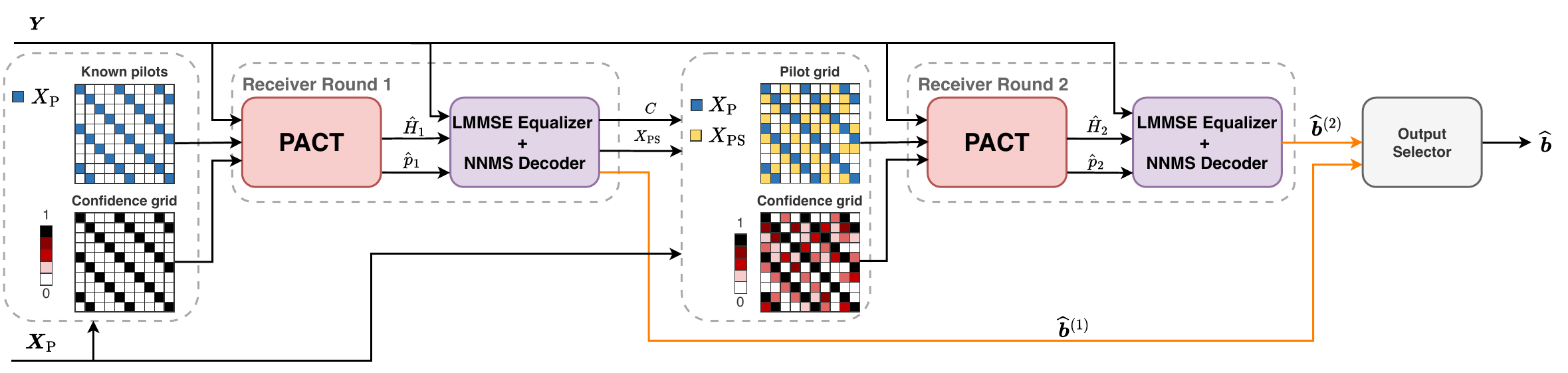}
    \caption{Two-round PACT receiver with rotating pilots and decoder-confidence-guided pseudo-pilot feedback.}
    \label{fig:pact_two_rounds}
\end{figure*}

\begin{figure*}[t]
    \centering
    \setlength{\abovecaptionskip}{3pt}
    \includegraphics[width=1\textwidth]{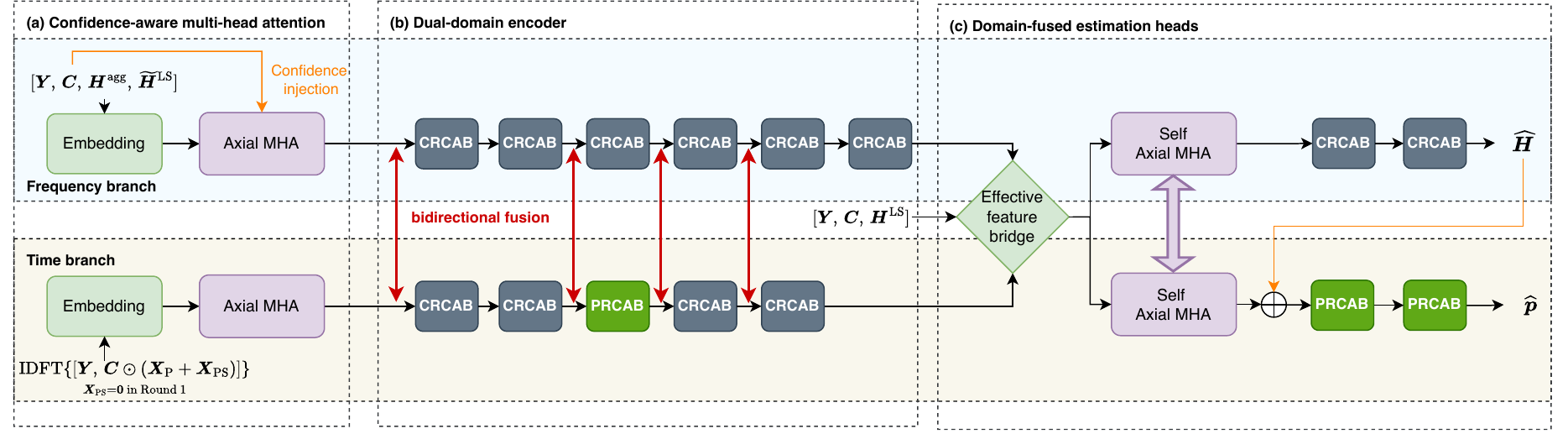}
    \caption{Architecture of the PACT estimator for joint channel and effective PN estimation.}
    \label{fig:pact_receiver}
\end{figure*}

Fig.~\ref{fig:pact_two_rounds} shows the PACT receiver,
which combines a PACT estimator shared across two rounds,
a banded LMMSE equalizer~\cite{rugini2006low},
and an NNMS decoder~\cite{nachmani2018deep}.
Known pilots follow a comb pattern with spacing
$D_{\mathrm p}$, cyclically shifted by one subcarrier
between consecutive OFDM symbols.
Round~1 estimates the channel and effective PN from
$\boldsymbol{Y}$ and $\boldsymbol{X}_{\mathrm P}$.
Decoder outputs provide hard pseudo-pilots
$\boldsymbol{X}_{\mathrm{PS}}$ and confidence weights
$\boldsymbol{C}$.
Round~2 reuses PACT with the same network parameters
as Round~1, using the original received grid
{$\boldsymbol{Y}$}, known pilots
{$\boldsymbol{X}_{\mathrm P}$} and pseudo-pilots
{$\boldsymbol{X}_{\mathrm{PS}}$}, and confidence
weights {$\boldsymbol{C}$} to refine both
estimates before another decoding pass.
A code-block-wise selector combines the two decoding
results into the final bit sequence $\widehat{\boldsymbol b}$.

\subsection{Pilot-Aware Complex Transformer}
\label{subsec:r1}

As shown in Fig.~\ref{fig:pact_receiver}, PACT uses interacting frequency- and time-domain branches, with frequency-domain inputs $\boldsymbol{Y}$, $\boldsymbol{C}$, $\widetilde{\boldsymbol{H}}^{\mathrm{LS}}$, and $\boldsymbol{H}^{\mathrm{agg}}$.
The LS grid $\boldsymbol{H}^{\mathrm{LS}}$ is estimated from
received and pilot grids, with additional PN compensation
using the Round~1 estimate in Round~2.
A small complex-valued convolutional network interpolates
$\boldsymbol{H}^{\mathrm{LS}}$ using confidence weights
$\boldsymbol{C}$ to obtain
$\widetilde{\boldsymbol{H}}^{\mathrm{LS}}$.
Approximate CPE compensation followed by confidence-weighted
aggregation of $\boldsymbol{H}^{\mathrm{LS}}$ across OFDM
symbols yields $\boldsymbol{H}^{\mathrm{agg}}$.
Time-domain inputs are IDFTs of $\boldsymbol{Y}$ and
the confidence-weighted pilot grid.
{After complex-valued convolutions embed the
input grids into learned feature channels, 
complex-valued axial multi-head attention (MHA) captures
dependencies along both grid axes.}
In the frequency-domain front end, subcarrier-axis MHA
incorporates confidence through gates
\begin{equation}
g_s(C)=1-\sigma(\gamma_s)(1-C),
\qquad s\in\{K,V\},
\label{eq:confidence_gating}
\end{equation}
and a learnable logit bias:
\begin{equation}
\begin{aligned}
A_{k,k'}
&=\operatorname{softmax}_{k'}\!\left(
\frac{g_K(C_{k',m})
\operatorname{Re}\{\mathbf{q}_k^H\mathbf{k}_{k'}\}}
{\sqrt{2d_h}}
+\alpha_{\mathrm c}C_{k',m}
\right),\\
\mathbf{o}_k
&=\sum_{k'}A_{k,k'}g_V(C_{k',m})\mathbf{v}_{k'}.
\end{aligned}
\label{eq:confidence_attention}
\end{equation}
{For each head at OFDM symbol $m$, the feature
at subcarrier $k$ is projected into the query $\mathbf{q}_k$,
while features at subcarriers $k'\in\{0,\ldots,N-1\}$ are
projected into keys $\mathbf{k}_{k'}$ and values
$\mathbf{v}_{k'}$, all of which are $d_h$-dimensional
complex vectors.
The softmax and summation run over $k'$, producing the
aggregated feature $\mathbf{o}_k$ at subcarrier $k$.
Outputs from all attention heads are concatenated
and fused through a learned complex-valued projection.}
Here, $\sigma$ denotes sigmoid, and $\alpha_{\mathrm c},\gamma_K,\gamma_V$ are learnable real scalars shared across heads.
Key gating rescales content scores without ensuring monotonic confidence weighting; no hard confidence mask is applied.

Complex residual channel attention blocks (CRCABs) capture local amplitude and phase correlations in frequency-selective, time-varying channels.
Time-branch phase-rotation residual channel attention blocks (PRCABs) introduce learnable phase rotations to facilitate PN compensation~\cite{wang2026crfcan}.
Multistage bidirectional fusion exchanges DFT/IDFT-aligned features through learnable projections to progressively refine both estimates.
The feature bridge fuses $[\boldsymbol{Y},\boldsymbol{H}^{\mathrm{LS}},\boldsymbol{C}]$ with both encoded streams and returns the result to both branches, preserving observation and reference-reliability information.
Subsequent axial self-attention captures long-range dependencies along both axes, sharing parameters across branches at matching depths.
Separate heads predict the frequency-domain channel $\widehat{\boldsymbol{H}}$ and time-domain effective PN $\widehat{\boldsymbol{p}}$, with $\widehat{\boldsymbol{H}}$ additionally conditioning PN recovery to distinguish propagation effects from PN distortion.

\subsection{Decoder-Confidence-Guided Refinement}
\label{subsec:r2}

In each round, time-domain PN compensation is followed by
banded LMMSE equalization to mitigate channel distortion
and ICI from neighboring subcarriers, after which the
equalized symbols are soft-demapped into log-likelihood ratios (LLRs) for NNMS decoding.
The equalizer uses a local linear fit to channel estimates
from three adjacent OFDM symbols to model intra-symbol
variation, retaining ICI coupling to the two nearest
subcarriers on either side.
Round~1 hard decisions for the transmitted coded bits are
remapped to QAM symbols to construct
$\boldsymbol{X}_{\mathrm{PS}}$, with data positions populated
and pilot positions set to zero.
Posterior LLRs are converted into bit probabilities,
whose products over the bit labels of each constellation
point yield a bit-factorized symbol posterior approximation
$\pi_{k,m}(a)$, where $a\in\mathcal{Q}$.
The continuous confidence is
\begin{equation}
c_{k,m}
=
\sum_{a\in\mathcal{Q}}\pi_{k,m}(a)
\left(
1-\frac{|X_{\mathrm{PS},k,m}-a|}{d_{\max}}
\right),
\label{eq:pseudo_pilot_confidence}
\end{equation}
where $d_{\max}=\max_{a,a'\in\mathcal{Q}}|a-a'|$.
This score measures the posterior-weighted geometric
closeness between the hard pseudo-pilot and possible
transmitted symbols, accounting for decoding uncertainty
and error distances rather than representing a probability
of symbol correctness.

The known pilot grid $\boldsymbol{X}_{\mathrm P}$ contains
known symbols at pilot positions and zeros at data positions.
Round~2 uses the augmented pilot grid
$\boldsymbol{X}_{\mathrm P}+\boldsymbol{X}_{\mathrm{PS}}$.
The confidence weight at position $(k,m)$ in Round~$r$
is $C_{k,m}^{(r)}$, with
\begin{equation}
\begin{aligned}
C_{k,m}^{(1)}
&=
\begin{cases}
1, & k\in\mathcal{P}_m,\\
0, & k\in\mathcal{D}_m,
\end{cases}
\quad
C_{k,m}^{(2)}
=
\begin{cases}
1, & k\in\mathcal{P}_m,\\
w_{\max}c_{k,m}, & k\in\mathcal{D}_m.
\end{cases}
\end{aligned}
\label{eq:round_confidence}
\end{equation}
Here, $w_{\max}$ is a constant that sets the maximum pseudo-pilot
weight, while all data references are retained without
hard confidence thresholds.
Finally, the output selector retains Round~1 code blocks
that satisfy the parity checks and uses Round~2 results
for the remaining blocks to form $\widehat{\boldsymbol b}$.

\subsection{Training Objective}
\label{subsec:training_objective}
PACT is first pretrained {without decoder feedback}
using only known pilots, supervised by the joint channel
and PN complex MSE loss
$\mathcal{L}_{\mathrm{pre}}=\ell_H^{(1)}+\ell_p^{(1)}$.
{In the following fine-tuning stage, half of the samples are used for training without decoder feedback, as in pretraining, while the other half are used for full two-round training with decoder feedback (pseudo-pilots and confidence) for Round~2, as described in Section~\ref{subsec:r2}.}
{The fine-tuning design is inspired by self-conditioning~\cite{chen2023analog}, and the fine-tuning objective is}
\begin{equation}
\mathcal{L}
= \ell_H^{(2)}+\ell_p^{(2)}
+\beta\,\ell_{\mathrm{bit}}^{(2)}
+\lambda\bigl(\ell_H^{(1)}+\ell_p^{(1)}\bigr),
\label{eq:training_objective}
\end{equation}
where $\ell_H^{(r)}$ and $\ell_p^{(r)}$ denote channel
and PN complex MSEs, respectively,
$\ell_{\mathrm{bit}}^{(2)}$ is {the binary cross-entropy
of the Round~2 output posteriors}, excluding filler bits, and $\lambda$ and
$\beta$ are loss weights.
{During fine-tuning, the PACT parameters are updated through supervision of both Round~1 and Round~2 outputs. Under this design, PACT learns to use pseudo-pilots and confidence feedback to further refine the estimates without degrading Round~1 performance.}
Training hyperparameters are specified in
Section~\ref{sec:experiments}.

\section{Experimental Setup}
\label{sec:experiments}

\subsection{Simulation and Training Setup}
\label{subsec:simulation_setup}

Table~\ref{tab:simulation_parameters} summarizes the main
settings of the training simulations, which use 3GPP
TDL channels~\cite{3gpp38901} and a 3GPP-based PN model whose PSD is shifted by $-10$~dB, with this spectral adjustment {based on} measured
synthesizer PN~\cite{liu2020synthesizer}.
Delay spreads are sampled from an equal mixture of InH
and UMi scenarios, with corresponding Rician K-factors
for LOS channels.
Training uses TDL-A/B/E with the RAN4 PN
model~\cite{3gpp38803}, whereas out-of-distribution
testing combines TDL-C/D with the RAN1 Set~1 and Set~2
PN models~\cite{3gpp38808}; the mixed-SCS datasets
contain equal proportions of 120, 240, and 480~kHz samples.
{Training includes both stationary and mobile users.}
\begin{table}[t]
\centering
\caption{Training Simulation and PACT Receiver Parameters}
\label{tab:simulation_parameters}
\small
\setlength{\tabcolsep}{3pt}
\begin{tabular}{@{}
p{\dimexpr0.5\columnwidth-\tabcolsep\relax}
p{\dimexpr0.5\columnwidth-\tabcolsep\relax}
@{}}
\toprule
Parameter & Value \\
\midrule
Carrier frequency $f_c$
    & 100~GHz \\
FFT size $N$
    & 1024 \\
OFDM symbols per frame $M$
    & 8 \\
Upsampling factor $U$
    & 8 \\
CP length $N_{\mathrm{cp}}$
    & 256 baseband samples \\
Data / pilot modulation
    & Gray 16QAM / QPSK \\
PACT pilot spacing $D_{\mathrm p}$
    & $8$ (overhead $12.5\%$) \\
Channel code
    & NR-LDPC, BG1, $Z=64$, RV0 \\
Effective code rate
    & $3/4$ \\
\bottomrule
\end{tabular}
\end{table}
{The NNMS decoder is separately trained on
AWGN-generated LLRs and remains frozen during PACT training.}
{The 48-channel PACT estimator is pretrained
for 80 epochs, then fine-tuned for 40 epochs.}
In both stages, PACT is trained over an $E_b/N_0$ range
of 20--30~dB using AdamW~\cite{loshchilov2019decoupled}
with an effective batch size of 8.
For pretraining and fine-tuning, {linear
warmup and cosine decay are applied.}
The loss weights in~\eqref{eq:training_objective}
are $\lambda=\beta=0.3$, with $w_{\max}=0.75$.
All network weights remain fixed during evaluation.

\setlength{\textfloatsep}{10pt plus 2pt minus 2pt}
\setlength{\floatsep}{6pt plus 2pt minus 2pt}

\subsection{Baselines and Performance Metrics}
\label{subsec:baselines}

Model-based baselines include IterativeLS~\cite{mohammadian2021deep},
which alternates channel and PN estimation
(23.44\% pilot overhead);
PT-RS~\cite{qi2018phase}, which estimates and compensates
CPE using phase-tracking reference signals (16.17\%);
and the Corvaja
receiver~\cite{corvaja2009joint}, which performs
decoder-aided two-pass channel and PN refinement (12.5\%).
Learning-based baselines include
DeepCascade~\cite{mohammadian2021deep}, whose OFDM variant
uses a three-stage neural architecture for channel estimation, PN compensation and detection (23.44\%);
DeepSRX~\cite{marasinghe2025phase}, which predicts bit LLRs
using a residual convolutional network, adapted from single-carrier transmission to CP-OFDM (12.5\%);
and CRFCAN~\cite{wang2026crfcan}, which jointly estimates
the channel and PN using a single-pass complex-valued
cross-domain network (12.5\%).

\begin{figure}[t]
    \centering
    \includegraphics[width=0.95\columnwidth,trim=0 12bp 0 0,clip]{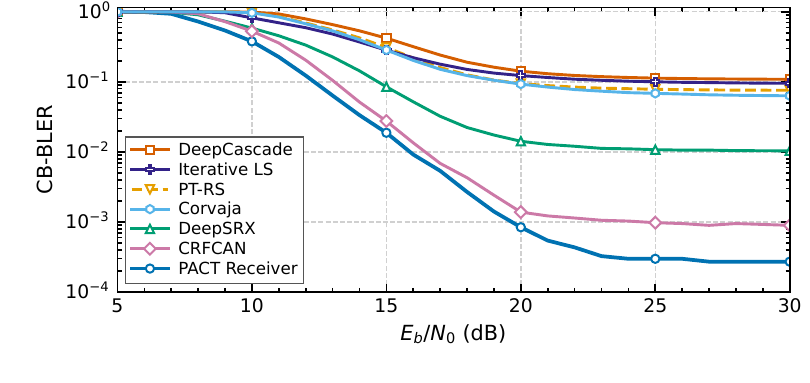}
    \caption{Code-block BLER in training-matched environments: TDL-A/B/E with RAN4 PN.}
    \label{fig:benchmark_ran4}
\end{figure}

\begin{figure}[t]
    \centering
    \includegraphics[width=0.95\columnwidth,trim=0 12bp 0 0,clip]{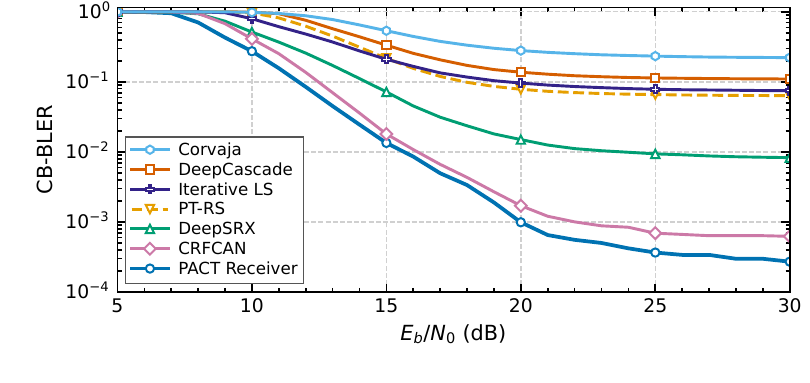}
    \caption{Code-block BLER in unseen environments: TDL-C/D with RAN1 Set~1/2 PN.}
    \label{fig:benchmark_ran1}
\end{figure}

\begin{figure}[t]
    \centering
    \includegraphics[width=0.95\columnwidth]{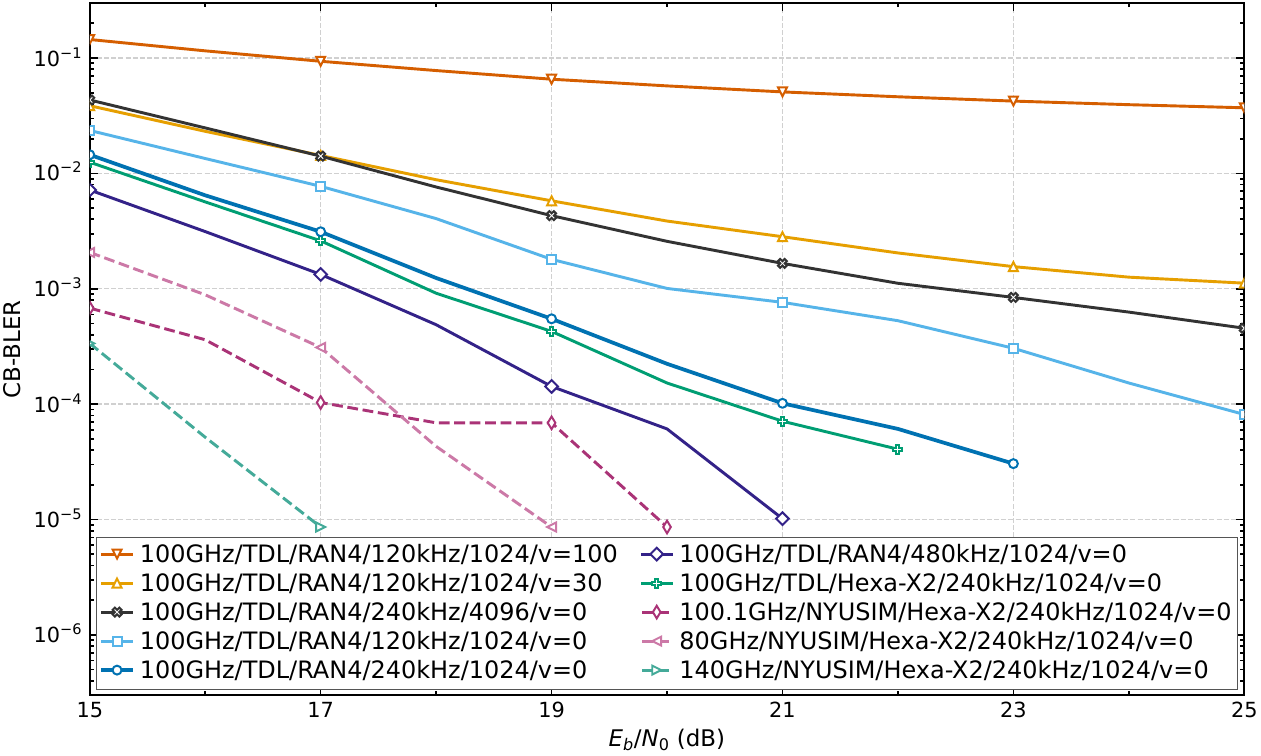}
    \caption{Generalization and applicability of the PACT receiver.}
    \label{fig:pact_generalization}
\end{figure}

\begin{figure}[t]
    \centering
    \includegraphics[width=0.95\columnwidth]{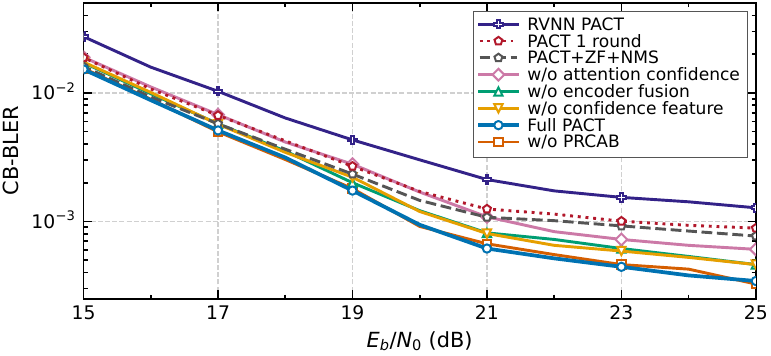}
    \caption{Code-block BLER of PACT receiver variants.}
    \label{fig:pact_ablation}
\end{figure}
Learning-based baselines are retrained on the same {
training-data distributions with approximately matched
numbers of updates per epoch.}
All baselines use NR-LDPC coding with a fixed normalized
min-sum (NMS) decoder (normalization factor $0.8$){;
single-tap zero-forcing (ZF) equalization is used except
in DeepSRX.}
Both the fixed NMS decoder and PACT's NNMS decoder
use 20 iterations per decoding pass.
Performance is evaluated by code-block error rate (BLER) and 
comparisons use information-bit $E_b/N_0$, accounting
for pilot and CP overheads.

\section{Results and Discussion}
\label{sec:results}

\subsection{Benchmark Performance}
\label{subsec:benchmark_results}

Fig.~\ref{fig:benchmark_ran4} evaluates TDL-A/B/E channels
with RAN4 PN, whereas Fig.~\ref{fig:benchmark_ran1} tests
unseen TDL-C/D channels with RAN1 Set~1/2 PN.
Both tests use realizations independent of training
and fixed network weights, with BLER computed by pooling
block-error and code-block counts across
120/240/480~kHz SCS.
The final outputs attain a BLER of $2.71\times10^{-4}$
in both test conditions at $E_b/N_0=30$~dB.
These values are over two orders of magnitude lower
than those of the two-pass Corvaja
receiver and approximately $70\%$ and $57\%$ lower
than those of CRFCAN, the strongest learning-based
baseline, demonstrating improved code-block recovery
under both test conditions.

\subsection{Generalization and Applicability}
\label{subsec:generalization}

Fig.~\ref{fig:pact_generalization} compares code-block BLER across configurations with frozen network weights.
Under stationary 100-GHz TDL-A/B/E channels with RAN4 PN, BLER at 20~dB decreases from $1.01\times10^{-3}$ to $2.24\times10^{-4}$ and $6.10\times10^{-5}$ as SCS increases from 120 to 240 and 480~kHz, demonstrating the shared model's applicability across the trained numerologies.
At 240~kHz, replacing RAN4 PN with Hexa-X Model~2~\cite{hexax2023d23} yields a BLER of $1.53\times10^{-4}$ at 20~dB.
NYUSIM channels~\cite{poddar2024nyusim} combined with Hexa-X PN at 80, 100.1, and 140~GHz yield observed BLERs no higher than $8.61\times10^{-6}$ at 20~dB, supporting applicability across channel/PN models and carrier frequencies.
Expanding FFT/CP from 1024/256 to 4096/1024 increases bandwidth and payload, with a higher BLER than the reference, reaching $4.55\times10^{-4}$ at 25~dB.
At 120~kHz, the 25-dB BLER increases from $8.14\times10^{-5}$ for stationary users to $1.12\times10^{-3}$ and $3.71\times10^{-2}$ at 30 and 100~km/h, respectively, reflecting increased Doppler sensitivity at small SCS.

\subsection{Ablation Study}
\label{subsec:ablation}

Fig.~\ref{fig:pact_ablation} pools code-block counts from TDL-A/B/C/D/E channels with RAN4 and RAN1 Set~1/2 PN.
{The following comparisons use \emph{Full PACT} at $E_b/N_0=25$~dB as the reference.}
\emph{RVNN PACT (w/o I/Q constraints)} retains the two-branch framework and receiver backend but uses a parameter-matched real-valued backbone without explicit I/Q coupling, Fourier-domain transformations or PRCABs, yielding $3.71$ times the BLER.
\emph{w/o PRCAB (Replaced by CRCAB)} replaces all PRCABs in both the time-domain encoder and PN head with CRCABs.
{The curves remain close, while this variant has 1.83M parameters compared with 1.69M for \emph{Full PACT}.}
\emph{w/o encoder fusion} removes bidirectional encoder fusion, raising BLER by approximately $34\%$.
\emph{w/o confidence feature} {sets the confidence input to zero, increasing BLER by $34\%$.}
\emph{w/o attention confidence} disables confidence injection in the frequency-branch encoder MHA, increasing BLER by $76\%$.
{These results support the effectiveness of PACT's physically motivated architectural choices.}
\emph{PACT+ZF+NMS} retains the \emph{Full PACT} weights and two-round feedback structure but replaces LMMSE equalization and NNMS decoding with single-tap ZF and fixed NMS in both rounds, yielding $2.24$ times the BLER.
\emph{PACT 1 round} performs a single PACT estimation pass followed by ZF equalization and NMS decoding, without confidence feedback and second-round refinement.
{Compared with \emph{PACT 1 round}, \emph{Full PACT} reduces BLER by approximately $61.2\%$.}
{Together, these comparisons support the benefits of confidence-guided feedback and the combined use of LMMSE equalization and NNMS decoding.}

\subsection{Computational Complexity}
\label{subsec:complexity}

\begin{table}[t]
    \centering
    \caption{Benchmark receiver computational cost.}
    \label{tab:receiver_cost_batch1}
    \footnotesize
    \setlength{\tabcolsep}{2pt}
    \renewcommand{\arraystretch}{1.00}
    \setlength{\aboverulesep}{0.5pt}
    \setlength{\belowrulesep}{0.5pt}
    \begin{tabular*}{\columnwidth}{@{\extracolsep{\fill}}lrrccl@{}}
        \toprule
        Method
        & \shortstack{Params\\(M)}
        & \shortstack{GFLOPs\\per frame}
        & \shortstack{Latency\\(ms/frame)}
        & \shortstack{GPU memory\\(MB)}
        & Device \\
        \midrule
        PACT, R1
        & 1.691 & 139.774 & 42.51 & 139.05 & H100 \\
        \textbf{PACT, R1+R2}
        & 1.691 & 279.551 & 85.79 & 144.10 & H100 \\
        DeepSRX
        & 0.676 & 11.005 & 9.40 & 76.60 & H100 \\
        DeepCascade
        & 1.018 & 19.091 & 9.84 & 60.39 & H100 \\
        CRFCAN
        & 2.957 & 95.677 & 31.41 & 98.19 & H100 \\
        \midrule
        PT-RS CPE
        & -- & 0.0295 & 70.54 & -- & CPU \\
        Iterative LS
        & -- & 0.0507 & 78.47 & -- & CPU \\
        Corvaja
        & -- & 0.1167 & 217.71 & -- & CPU \\
        \bottomrule
    \end{tabular*}
\end{table}
Table~\ref{tab:receiver_cost_batch1} reports computational
costs measured on an NVIDIA H100 80GB HBM3 GPU for
learning-based receivers and an Intel Xeon Gold 6448Y CPU
for model-based baselines.
With shared PACT parameters, two-round processing increases
peak GPU memory only from 139.05 to 144.10~MB, while latency
rises from 42.51 to 85.79~ms/frame.
Compared with CRFCAN, PACT uses fewer parameters but requires
more computation and higher latency, reflecting a tradeoff
between reception performance and computational cost.
CPU and GPU latencies are provided for reference and are
not directly comparable.

\section{Conclusion}
\label{sec:conclusion}

The PACT receiver integrates decoder-confidence-guided
two-round joint channel and PN estimation with LMMSE
equalization and NNMS decoding for coded sub-THz CP-OFDM.
Benchmark results show approximately $70\%$ and $57\%$
lower BLER than the strongest learning-based baseline
in the two settings at $E_b/N_0=30$~dB, while ablation
results show that the full PACT receiver reduces BLER
by $61.2\%$ relative to PACT 1 round 
at $E_b/N_0=25$~dB.
Frozen-weight tests further demonstrate applicability
across the evaluated configurations, {including different
subcarrier spacings, carrier frequencies, and channel
and PN models.}

\bibliographystyle{IEEEtran}
\bibliography{references}

\end{document}